%% file: main.tex
\documentclass[10pt,twocolumn,letterpaper]{article}

\usepackage{cvpr}              

\input{preamble}

\definecolor{cvprblue}{rgb}{0.21,0.49,0.74}
\usepackage[pagebackref,breaklinks,colorlinks,allcolors=cvprblue]{hyperref}

\def\paperID{12687} 
\def\confName{CVPR}
\def\confYear{2026}

\title{MotionScale: Reconstructing Appearance, Geometry, and Motion of Dynamic Scenes with Scalable 4D Gaussian Splatting}

\author{Haoran Zhou \quad Gim Hee Lee\\
Department of Computer Science, National University of Singapore\\
{\texttt{\small haoran.zhou@u.nus.edu, gimhee.lee@nus.edu.sg }}\\
}

\begin{document}
\maketitle

\begin{strip}
\centering
\includegraphics[width=0.98\textwidth]{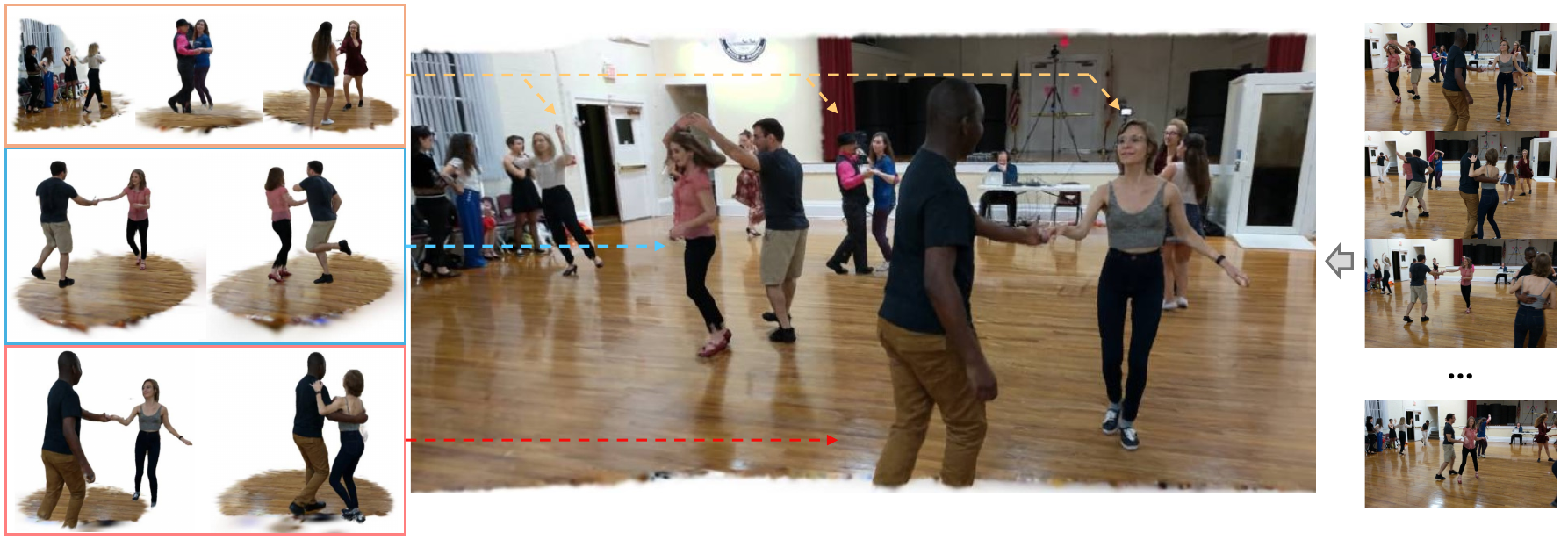}
\captionof{figure}{Visualization of a reconstructed dynamic scene and extracted moving objects. Given a single monocular video as input, \textbf{MotionScale} reconstructs a 4D scene representation that effectively captures photorealistic appearance, accurate 3D geometry, diverse human motion. Refer to the supplementary material for video results and additional examples.}
\label{fig:teaser}
\end{strip}



\input{sec/0_abstract}    
\input{sec/1_intro}

\input{sec/2_related_work}
\input{sec/3_method}
\input{sec/4_experiments}

\vspace{-10pt}
\paragraph{Acknowledgment.} This research / project is supported by the National Research Foundation (NRF) Singapore, under
its NRF-Investigatorship Programme (Award ID. NRF-NRFI09-0008).

{
    \small
    \bibliographystyle{ieeenat_fullname}
    \bibliography{main}
}


\end{document}

%% file: preamble.tex
\usepackage{multirow}
\usepackage{caption}
\usepackage{subcaption}
\usepackage{graphicx}
\usepackage{amsmath}
\usepackage{soul}
\usepackage{enumitem}
\usepackage{float}
\usepackage{tikz}
\usepackage{capt-of}
\usepackage{cuted}

%% file: sec/0_abstract.tex
\begin{abstract}
Realistic reconstruction of dynamic 4D scenes from monocular videos is essential for understanding the physical world.
Despite recent progress in neural rendering, existing methods often struggle to recover accurate 3D geometry and temporally consistent motion in complex environments.
To address these challenges, we propose MotionScale, a 4D Gaussian Splatting framework that scales efficiently to large scenes and extended sequences while maintaining high-fidelity structural and motion coherence.
At the core of our approach is a scalable motion field parameterized by cluster-centric basis transformations that adaptively expand to capture diverse and evolving motion patterns.
To ensure robust reconstruction over long durations, we introduce a progressive optimization strategy comprising two decoupled propagation stages: 1) A background extension stage that adapts to newly visible regions, refines camera poses, and explicitly models transient shadows; 2) A foreground propagation stage that enforces motion consistency through a specialized three-stage refinement process.
Extensive experiments on challenging real-world benchmarks demonstrate that MotionScale significantly outperforms state-of-the-art methods in both reconstruction quality and temporal stability.
Project page: {\small \url{https://hrzhou2.github.io/motion-scale-web/}}.

\end{abstract}

%% file: sec/1_intro.tex
\section{Introduction}
\label{sec:intro}
Understanding and reconstructing dynamic 4D scenes is a pivotal challenge in computer vision, essential for enabling machines to perceive and interact with the physical world. 
Recently, the emergence of geometric foundation models has revolutionized the recovery of 3D structures from unconstrained images. Frameworks such as DUST3R~\cite{wang2024dust3r} and VGGT~\cite{wang2025vggt} have demonstrated remarkable capability in estimating dense correspondences and inferring underlying 3D geometry from sparse or monocular observations. These models provide powerful geometric and motion priors that generalize across diverse environments, underpinning a wide range of downstream applications, including autonomous driving and motion forecasting~\cite{shi2025motion,ettinger2021large}, AR/VR content creation~\cite{schafer2022survey} and immersive telepresence~\cite{hilty2020review}.


Despite the strong priors provided by these 2D models, high-fidelity 4D reconstruction, which aims to recover synchronized appearance, geometry, and motion across dynamic environments, remains an open challenge.
Existing approaches~\cite{pumarola2021d,park2021nerfies,park2021hypernerf,chen2022tensorf,fridovich2023k,wu20244d,luiten2024dynamic,yang2023gs4d} based on Neural Radiance Fields (NeRFs)~\cite{mildenhall2021nerf} and 3D Gaussian Splatting (3DGS)~\cite{kerbl20233d} have achieved remarkable success in photorealistic synthesis for static or mildly dynamic settings with dense multi-view supervision.
However, scaling these representations to unconstrained, in-the-wild environments remains a significant bottleneck.
More recent works~\cite{wang2025shape,huang2024sc,park2025splinegs,wang2025gflow,wu20254d} have begun to integrate efficient 4D Gaussian Splatting with 2D geometric and motion priors to reconstruct scenes from casual monocular video.
While these methods yield plausible view synthesis from observed viewpoints, they frequently suffer from geometric distortion and lack temporal coherence across extended sequences. 
We identify two primary bottlenecks hindering current state-of-the-art approaches: 
1) \textit{Under-constrained Geometry}: Supervision predominantly relies on view-dependent appearance signals, which lack the capacity to enforce strict 3D structural consistency for dynamic objects.
2) \textit{Accumulated Temporal Drift}: Motion models often rely on 2D tracking priors that lack 3D awareness. Over long-duration sequences, the representation inevitably accumulate errors, resulting in geometric collapse and inconsistent motion trajectories.

To address these challenges, we propose MotionScale, a scalable 4D Gaussian Splatting framework designed for the high-fidelity reconstruction of large-scale dynamic scenes. 
Unlike prior methods that rely on global deformation fields or fixed-capacity architectures, MotionScale introduces a cluster-centric motion representation that adaptively expands to capture diverse motion patterns across space and time. 
This design allows our motion field to scale with scene complexity while maintaining both spatio-temporal consistency and computational efficiency.
Furthermore, we develop a progressive optimization strategy that scales seamlessly to unseen frames by effectively incorporating newly visible regions and refining motion trajectories across extended sequences. 
We evaluate MotionScale on several challenging real-world benchmarks, demonstrating that it achieves state-of-the-art reconstruction quality and motion consistency, significantly outperforming existing 4D Gaussian Splatting methods.
Our main contributions are:
\begin{itemize}
    \item A scalable 4D Gaussian Splatting framework, MotionScale, that effectively reconstructs large-scale dynamic scenes with accurate geometry, photorealistic appearance, and coherent motion across long sequences.

    \item A cluster-centric motion field that adaptively expands to model complex and unbounded motion patterns.

    \item A progressive optimization strategy featuring decoupled foreground and background propagation stages that ensure stable convergence and temporal coherence.
\end{itemize}


%% file: sec/2_related_work.tex
\section{Related Work}
\label{sec:related_work}

\noindent \textbf{2D vision models for 3D scene understanding.}
Recent progress in 2D foundation models has greatly advanced 3D scene understanding, which provides strong per-frame priors for tasks such as depth estimation, point tracking, and semantic segmentation~\cite{kirillov2023segment,ravi2024sam}.
In monocular depth estimation, large-scale pre-trained models now deliver high-quality per-frame depth with strong zero-shot generalization~\cite{bhat2023zoedepth,ranftl2020towards,yang2024depth,yang2024depthv2,chen2025video}.
Among these methods, Depth Anything V2~\cite{yang2024depthv2} is a large-scale depth model with up to 1.3B parameters. Trained on 595K synthetic labeled images and over 62M unlabeled real images, it produces significantly finer and more robust depth predictions than its predecessor~\cite{yang2024depth}.
However, although Depth Anything models produce high-quality depth maps for individual images, they do not enforce consistent scene geometry across views. This limitation has motivated methods that recover metric-scale 3D geometry from single-view observations~\cite{wang2025moge,piccinelli2024unidepth,piccinelli2025unidepthv2,lu2025align3r}.
Specifically, MoGe~\cite{wang2025moge} addresses single-view geometry by directly predicting a dense 3D point map from a single image, which yields accurate and fine-grained reconstruction of scene structure.

For correspondence and motion, foundation models now enable tracking of arbitrary points throughout the entire video. 
This goes beyond traditional optical flow methods, which are limited to short-term frame-to-frame correspondences~\cite{dosovitskiy2015flownet,ilg2017flownet,sun2018pwc,teed2020raft,huang2022flowformer}.
The Tracking Any Point (TAP) paradigm was first introduced by TAP-Vid~\cite{doersch2022tap}, which established a benchmark and proposed TAP-Net as a baseline model for point tracking.
Building on this foundation, many methods have extended the TAP paradigm~\cite{zheng2023pointodyssey,doersch2023tapir,karaev2024cotracker,karaev2025cotracker3,doersch2024bootstap}. 
Among these methods, TAPIR~\cite{doersch2023tapir} improves point tracking with a two-stage network that matches and refines query point positions across video frames.
Another line of work is CoTracker~\cite{karaev2024cotracker}, which introduces a transformer-based architecture for joint tracking of dense point sets over long video sequences and uses proxy tokens to improve efficiency. 
CoTracker3~\cite{karaev2025cotracker3} further simplifies the architecture and adopts a semi-supervised training strategy on real unlabeled videos. It reaches state-of-the-art performance while using substantially less training data.
Furthermore, SpatialTracker~\cite{xiao2024spatialtracker} extends point tracking into 3D  
and jointly estimates depth and motion to maintain temporally consistent point trajectories in dynamic scenes.
Beyond single tasks, joint feed-forward geometry models such as VGGT, $\pi^3$ and MapAnything infer camera poses, depth or point-maps and multi-view structure directly from images \cite{wang2025vggt,wang2025pi3,keetha2025mapanything}.

\medskip
\noindent \textbf{4D Scene Reconstruction.} 
Reconstructing dynamic 3D scenes and generating novel views (NVS) is an active research area in computer vision.
Neural Radiance Fields (NeRF)~\cite{mildenhall2021nerf} pioneered high-fidelity view synthesis for static scenes, and subsequent dynamic NeRFs~\cite{pumarola2021d,park2021nerfies,park2021hypernerf,li2021neural,chen2022tensorf,fridovich2023k,muller2022instant} extend this framework with canonical-space fields and time-varying deformations to model dynamic scenes.
However, NeRF-based approaches require intensive volumetric optimization and are computationally expensive and inefficient for real-time rendering. This limitation has motivated point-based radiance representations that improve efficiency.
3D Gaussian Splatting (3DGS)~\cite{kerbl20233d} marked a major milestone by representing scenes as a set of 3D Gaussians and achieving real-time photorealistic rendering.
Recent work builds on the expressive 3DGS representation and extends it to dynamic scenes~\cite{wu20244d,luiten2024dynamic,yang2023gs4d,wang2025shape,huang2024sc,park2025splinegs,wang2025gflow,wu20254d} with the capability of modeling complex geometry and motion from casual in-the-wild videos.
These approaches significantly improve training and rendering efficiency as well as visual quality for dynamic novel view synthesis, yet they still face challenges in recovering accurate geometry, handling complex motion, and scaling to large-scale scenes and long video sequences.

%% file: sec/3_method.tex
\begin{figure*}
    \centering
    \includegraphics[width=0.99\linewidth]{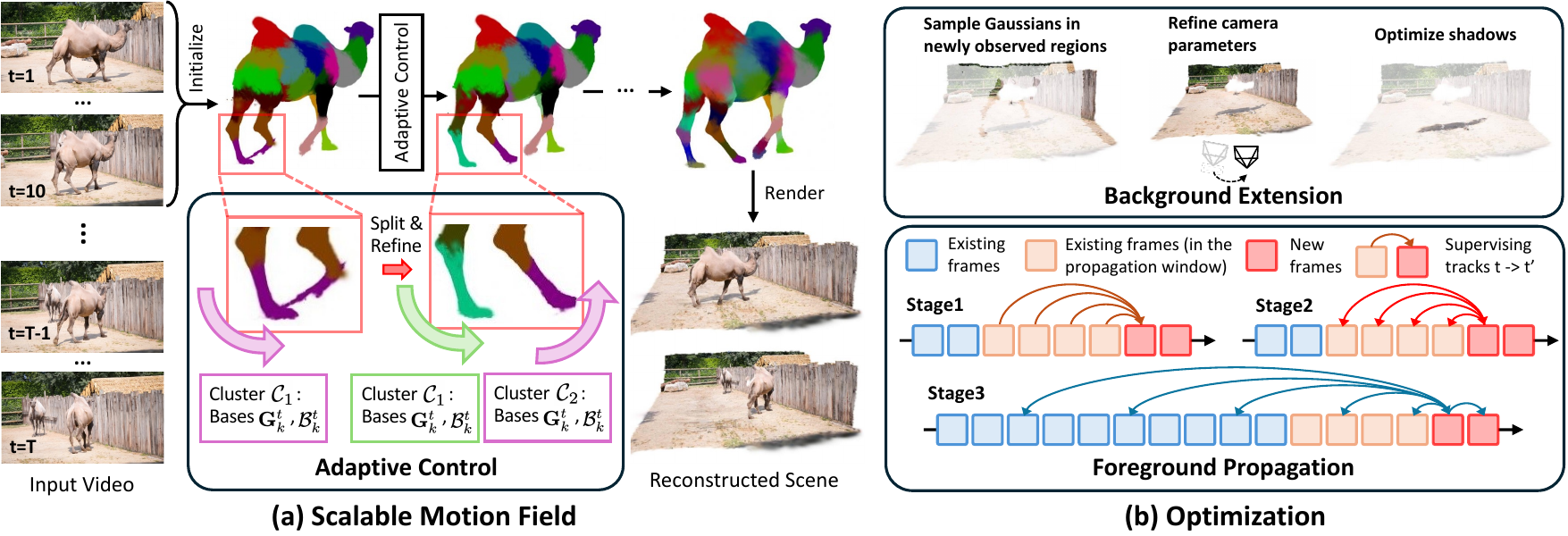}
    \caption{\textbf{Overview of MotionScale.} Our method adopts a scalable motion field that progressively captures object motions through an adaptive control mechanism, enabling efficient splitting and refinement of motion components. For optimization, the background is updated through region sampling, camera refinement, and shadow handling, while the foreground propagation employs a three-stage refinement to propagate motion across long temporal windows for consistent 4D reconstruction.}
    \label{fig:pipeline}
\end{figure*}

\section{Method}
\label{sec:method}


\noindent \textbf{Overview.} Fig.~\ref{fig:pipeline} provides an overview of MotionScale. 
Building on 3D Gaussian Splatting (3DGS) (\cf~Sec.~\ref{sec:method:3dgs}), we represent dynamic scenes as a set of canonical 3D Gaussians governed by a scalable motion field that models complex and diverse motion patterns while maintaining spatio-temporal consistency (\cf~Sec.~\ref{sec:method:motion_field}).
The motion field leverages cluster-based basis transformations to parameterize the dynamics of local regions and employs an adaptive control strategy that autonomously expands or prunes clusters to achieve scalability.
MotionScale adopts a progressive optimization strategy that jointly refines appearance, geometry, and motion while scaling seamlessly to long video sequences (\cf~Sec.~\ref{sec:method:optim}).
The optimization proceeds via temporal propagation across  frames, where a \textit{background extension} stage refines unseen regions, camera poses, and shadows, while a \textit{foreground propagation} stage employs a three-stage refinement to enforce long-term motion coherence.

\subsection{Preliminary: 3D Gaussian Splatting}
\label{sec:method:3dgs}
We build upon 3D Gaussian Splatting (3DGS)~\cite{kerbl20233d} as the foundational representation of our method.
Standard 3DGS models a static scene as a collection of $N$ Gaussian primitives defined in a canonical coordinate frame, $g_i^0 = \{\boldsymbol{\mu}_i^0,\mathbf{R}_i^0,\mathbf{s}_i,o_i,\mathbf{c}_i\}$, where $i=1,\dots,N$.
Here, $\boldsymbol{\mu}_i^0 \in \mathbb{R}^3$ and $\mathbf{R}_i^0 \in \mathrm{SO}(3)$ represent the center position and orientation of the $i$-th Gaussian in 3D space. The attributes $\mathbf{s}_i \in \mathbb{R}^3$, $o_i \in \mathbb{R}$, and $\mathbf{c}_i \in \mathbb{R}^3$ denote its scale, opacity, and color, respectively.
During rendering, for a given pixel $\mathbf{p}$ in the image view $\mathbf{I}$ with extrinsic matrix $\mathbf{E}$ and intrinsic matrix $\mathbf{K}$, the pixel color $\mathbf{I}(\mathbf{p})$ is obtained by alpha-blending the colors of all intersected 3D Gaussians as:
\begin{equation}
    \mathbf{I}(\mathbf{p}) = \sum_{i \in H(\mathbf{p})} \mathbf{c}_i \alpha_i \prod_{j=1}^{i-1}(1-\alpha_j),
    \label{eq:3dgs_render}
\end{equation}
where $H(\mathbf{p})$ denotes the set of Gaussians projected at pixel $\mathbf{p}$. Each Gaussian contributes to the final color according to its opacity term, $\alpha_i = o_i \cdot \exp(-\frac{1}{2}(\mathbf{p}-\boldsymbol{\mu}^{\text{2d}}_i)^\top (\boldsymbol{\Sigma}^{\text{2d}}_i)^{-1} (\mathbf{p}-\boldsymbol{\mu}^{\text{2d}}_i))$, where $\boldsymbol{\mu}^{\text{2d}}_i$ and $\boldsymbol{\Sigma}^{\text{2d}}_i$ represent projected 2D mean and covariance on the image plane.
Building on this, we represent a dynamic scene as a set of canonical 3D Gaussians $\{g_i^0\}$ and a time-dependent motion field that maps each Gaussian to its state at time $t$.

\subsection{Scalable Motion Field}
\label{sec:method:motion_field}
A fundamental challenge in large-scale dynamic reconstruction lies in designing a motion representation that is both expressive and scalable. 
Conventional deformation fields often rely on fixed-capacity architectures, such as global MLPs or a predefined set of temporal bases, which struggle to resolve localized, diverse object motions.
In contrast, we propose a \textit{cluster-based motion field} that enables an adaptive and scalable allocation of model capacity.
Specifically, the dynamic Gaussians $\mathcal{G}_d = \{g_i\}_{i=1}^{N_d}$ are partitioned into $K$ disjoint clusters $\{\mathcal{C}_k\}_{k=1}^{K}$. 
For each cluster $\mathcal{C}_k$, we define a hierarchical motion model consisting of a global transformation and a set of local refinement bases. 
The global transformation $\mathbf{G}_k^t = [\mathbf{R}_{k,g}^t \mid \mathbf{t}_{k,g}^t] \in \mathrm{SE}(3)$ captures the global rigid movement of the entire cluster. 
To model localized non-rigid deformations, we define $B$ fine-grained basis transformations $\mathcal{B}_k^t = \{(\mathbf{r}_{k,b}^t, \mathbf{t}_{k,b}^t)\}_{b=1}^B$, where $\mathbf{r}_{k,b}^t$ is a continuous rotation representation (e.g., 6D~\cite{zhou2019continuity}) and $\mathbf{t}_{k,b}^t \in \mathbb{R}^3$. 
Each Gaussian $g_i \in \mathcal{C}_k$ is assigned a learnable coefficient vector $\mathbf{w}_i = [w_{i,1}, \dots, w_{i,B}]^\top$, where $\sum_{b=1}^B w_{i,b} = 1$. 
The local transformation $\mathbf{L}_i^t = [\mathbf{R}_{i, \ell}^t \mid \mathbf{t}_{i, \ell}^t]$ is computed by blending the cluster bases:
\begin{equation}
\mathbf{R}_{i, \ell}^t = \mathcal{R} \left( \sum_{b=1}^B w_{i,b} \mathbf{r}_{k,b}^t \right), \quad \mathbf{t}_{i, \ell}^t = \sum_{b=1}^B w_{i,b} \mathbf{t}_{k,b}^t,
\end{equation}
where $\mathcal{R}(\cdot)$ denotes the mapping to $\mathrm{SO}(3)$.
The final state of Gaussian $g_i$ at time $t$ is the composition of global and local transformations applied to the canonical state $(\boldsymbol{\mu}_i^0, \mathbf{R}_i^0)$:
\begin{equation}
\boldsymbol{\mu}_i^t = \mathbf{R}_{k,g}^t ( \mathbf{R}_{i, \ell}^t \boldsymbol{\mu}_i^0 + \mathbf{t}_{i, \ell}^t ) + \mathbf{t}_{k,g}^t, \mathbf{R}_i^t = \mathbf{R}_{k,g}^t \mathbf{R}_{i, \ell}^t \mathbf{R}_i^0.
\end{equation}

\noindent\textbf{Adaptive control.}
To dynamically adjust representation capacity, we introduce an adaptive control mechanism inspired by the densification strategy of 3D Gaussians~\cite{kerbl20233d}. 
This scheme splits or culls clusters based on their fidelity in representing local dynamics.
Since each cluster is designed to model a roughly rigid body segment, any significant non-rigid or inconsistent motion within a cluster indicates that the current representation is insufficient.
To resolve this, as illustrated in Fig.~\ref{fig:pipeline}(a), we partition the cluster's Gaussians $\{g_i \in \mathcal{C}_k\}$ into two groups exhibiting distinct motion patterns (detailed in Sec.~\ref{sec:method:optim}).
To ensure optimization stability, we duplicate the original motion parameters for both new clusters.
This initialization allows the two clusters to diverge in parameter space while fully preserving the original motion state at the moment of splitting, ultimately enabling the motion field to capture finer, localized variations.

\smallskip
\noindent\textbf{Efficiency and scalability.}
The cluster-based motion field is designed for both computational efficiency and architectural scalability.
Because each Gaussian is influenced by a single cluster, the computational cost remains nearly constant even as the motion field expands.
Furthermore, the representation is highly memory-efficient, as each cluster maintains a compact parameter set that scales marginally relative to the total Gaussian count.
Beyond efficiency, this architectural flexibility also ensures high expressiveness. 
As illustrated in Fig.~\ref{fig:pipeline}(a), the splitting process isolates distinct or non-rigid movements into independent clusters, allowing the motion field to resolve previously ambiguous dynamics.
By progressively optimizing these new clusters, our method can capture intricate details and cover all meaningful parts of an object (\cf Fig.~\ref{fig:pipeline}(a), top row).
Consequently, this adaptive control enables our framework to scale seamlessly to larger scenes and longer sequences while maintaining both coherent geometry and precise motion tracking.


\subsection{Optimization Strategy}
\label{sec:method:optim}
Given a video sequence of $T$ images $\{I_t\}_{t=1}^T$ with unknown camera parameters, our goal is to reconstruct a 4D scene representation that captures high-quality geometry, photorealistic appearance, and accurate motion. 
To provide a robust initialization for both geometry and dynamics, we leverage pre-trained 2D models to extract a suite of priors, including monocular depth maps $\{\mathbf{D}_t\}_{t=1}^T$, foreground masks $\{\mathbf{M}_t\}_{t=1}^T$, and dense 2D point tracks $\{\mathbf{U}_t\}_{t=1}^T$. 
Additionally, we estimate initial camera poses using vision-based geometry frameworks~\cite{wang2025pi3}, which are jointly refined with the scene representation during optimization.


\smallskip
\noindent\textbf{Initialization.}
We represent the scene as a combination of a static background and a dynamic foreground that are jointly rendered to produce the final image~\cite{huang2024sc,wang2025shape}. 
The optimization begins on an initial temporal window $\{I_t\}_{t=1}^{T_{\text{init}}}$. 
To initialize the background, we back-project 3D points from the monocular depth maps $\{\mathbf{D}_t\}$ to create an initial 3DGS point cloud. 
For the dynamic motion field, we sample 3D trajectories from the 2D tracks $\{\mathbf{U}_t\}$ and the corresponding depth maps.
We then apply K-means clustering to these 3D points in the canonical frame to define the initial spatial extent of the $K$ clusters. 
The global transformations $\mathbf{G}_k^t$ are then initialized via Procrustes analysis to estimate the rigid transformations between point clouds across the initial frames, while local refinement bases $\mathcal{B}_k^t$ are set to identity.

\smallskip
\noindent\textbf{Progressive optimization.}
To handle long video sequences efficiently while maintaining temporal coherence, we adopt a progressive optimization strategy (\cf Fig.~\ref{fig:pipeline}(b)). 
Once the initial window is optimized, we incrementally incorporate a subset of $T_{\text{new}}$ additional frames. 
This iterative process consists of two primary steps: background extension and foreground adaptation.


\smallskip
\noindent\textbf{Background extension.}
To incorporate a batch of $T_{\text{new}}$ additional frames, we first extend the static background reconstruction. 
Since new views often reveal previously occluded regions or areas beyond the initial image boundaries, we identify and systematically populate these unobserved regions to maintain background completeness.
Specifically, we project all existing background Gaussians onto the new image planes, marking pixels without coverage as unobserved areas that require new samples.
In these regions, we initialize additional Gaussians by sampling 3D points from the monocular depth maps $\{\mathbf{D}_t\}$. 
We then perform a targeted optimization on the background pixels of the new frames to refine the added geometry and ensure photometric consistency. 
Simultaneously, we jointly refine the camera extrinsics via an end-to-end gradient-based approach. 
This lightweight refinement corrects sub-pixel misalignments in the initial poses~\cite{wang2025pi3} through direct photometric loss, providing a simple yet effective alternative to explicit SLAM-based tracking or global bundle adjustment.

\smallskip
\noindent\textbf{Foreground propagation.}
Following the background update, we extend the motion field to the newly introduced frames. 
The objective is to estimate the global transformations $\mathbf{G}_k^t$ and the local bases $\mathcal{B}_k^t$ that parameterize the scene dynamics at these new timestamps.
To ensure temporal continuity, we initialize the motion bases for the $T_{\text{new}}$ frames based on the optimized transformations from the most recent frames.
The estimation is supervised using 2D tracking priors. 
For each pixel $p$ in a query frame $t$, we compute its 3D trajectory at a target time $t'$ by accumulating the 3D positions of the Gaussians via alpha-blending:
\begin{equation}
  \mathbf{X}_{t \rightarrow t'}(\mathbf{p}) = \sum_{i \in H(\mathbf{p})} \boldsymbol{\mu}_i^{t'} \alpha_i \prod_{j=1}^{i-1}(1-\alpha_j), 
\end{equation}
where $H(\mathbf{p})$ denotes the set of Gaussians projected at pixel $p$, and $\boldsymbol{\mu}_i^{t'}$ is the position of Gaussian $g_i$ at time $t'$, derived from its associated cluster transformations $\mathcal{T}_k^{t'}$.
The 3D position $\mathbf{X}_{t \rightarrow t'}(\mathbf{p})$, representing the estimated world coordinates for pixel $p$ at time $t'$, is then projected onto the image plane to obtain the estimated 2D track $\hat{\mathbf{U}}_{t \rightarrow t'}(\mathbf{p})$ and depth $\hat{\mathbf{D}}_{t \rightarrow t'}(\mathbf{p})$.
The motion field is then optimized by minimizing the tracking and depth consistency losses:
\begin{equation}
  L_{\text{track}} = \frac{1}{|I_t|} \sum_{p\in I_t} \| \hat{\mathbf{U}}_{t \rightarrow t'}(\mathbf{p}) - \mathbf{U}_{t \rightarrow t'}(\mathbf{p})\|,
  \label{eq:trackloss}
\end{equation}
\begin{equation}
  L_{\text{depth}} = \frac{1}{|I_t|} \sum_{p\in I_t} \| \hat{\mathbf{D}}_{t \rightarrow t'}(\mathbf{p}) - \mathbf{D}_{t \rightarrow t'}(\mathbf{p})\|.
  \label{eq:depthloss}
\end{equation}
where $\mathbf{U}_{t \rightarrow t'}(\mathbf{p})$ is the 2D track prior, denoting the target location in frame $t'$ for a pixel $p$ from frame $t$, and $\mathbf{D}_{t \rightarrow t'}(\mathbf{p}) = \mathbf{D}_{t'}(\mathbf{U}_{t \rightarrow t'}(\mathbf{p}))$ represents the monocular depth prior evaluated at that tracked location.

To mitigate noise in the 2D tracking priors and ensure temporal consistency, we optimize the motion field through a \textit{three-stage refinement process} (\cf Fig.~\ref{fig:pipeline}(b)):

\begin{enumerate}[label=\arabic*., leftmargin=15pt, itemsep=4pt, topsep=4pt]
\item \textit{Initial Alignment.}
We first define a short propagation window (denoted as orange frames in Fig.~\ref{fig:pipeline}(b)) covering the $T_{\text{prop}}$ most recent frames. To avoid polluting the well-optimized previous frames with unaligned new data, we employ a one-directional tracking loss using $\mathbf{U}_{t \rightarrow t'}(\mathbf{p})$, where $t$ belongs to the established frames and $t'$ represents the new frames. In this stage, we optimize only the motion bases for the new frames while freezing all other scene parameters, including Gaussian attributes and existing transformations.

\item \textit{Short-term Consistency.} 
Once the new frames are roughly aligned, we transition to a \textit{bi-directional tracking loss} computed over arbitrary frame pairs $(t, t')$ within the propagation window. This stage enforces local temporal consistency and allows gradients to flow back into recent frames, refining their states with information from the new views. 
Stages 1 and 2 are repeated iteratively as new frames are incorporated. 
This stage remains efficient by limiting the optimization to the motion field and leveraging the sparsity of the 2D tracking priors.

\item \textit{Long-term Refinement.} 
After a sufficient number of frames have been integrated, we perform long-term motion refinement by sampling frame pairs $(t, t')$ across the entire optimized sequence. 
This global supervision resolves accumulated drift and occlusions. 
At this stage, we jointly optimize all parameters and allow for Gaussian densification. 
This joint optimization objective combines the previously defined tracking and depth losses with a photometric (RGB) loss and as-rigid-as-possible (ARAP) regularization terms.

\end{enumerate}

\noindent\textbf{Adaptive control.}
During the long-term refinement in Stage 3, we trigger the adaptive control mechanism to dynamically update the cluster topology. 
We first identify and partition clusters that fail to accurately represent the coherent motion of their associated Gaussians. 
Specifically, for each cluster, we extract the 3D trajectories of its Gaussians over the propagation window to serve as feature descriptors for HDBSCAN clustering. 
Within the resulting density-based sub-clusters, we apply Agglomerative Clustering to isolate two candidate groups and compute the distance between their centroids. 
If this distance exceeds a predefined threshold, the original cluster is identified as motion-inconsistent. 
As illustrated in Fig.~\ref{fig:pipeline}(a), this approach effectively isolates Gaussians with divergent dynamics. 
Following the splitting protocol in Sec.~\ref{sec:method:motion_field}, new clusters and motion bases are initialized to accommodate these new dynamics. 
Finally, we prune undersized clusters by eliminating their associated Gaussians and removing the corresponding motion field entries, followed by a remapping of the remaining cluster indices to ensure a compact representation.


\smallskip
\noindent\textbf{Shadow Gaussians.}
To realistically reconstruct the transient lighting effects caused by moving objects, we introduce a dedicated set of ``shadow Gaussians'' within the background representation. Unlike static background primitives, these Gaussians are coupled with a dynamic motion field, allowing them to track with the objects casting the shadows. Recognizing that shadows lack well-defined 3D geometry and that their masks are often temporally inconsistent, we deliberately omit geometric and motion supervision for these primitives. Instead, shadow Gaussians are optimized primarily via the photometric (RGB) loss, complemented by a segmentation constraint to prevent spatial overlap with the foreground object. Initialized from coarse shadow masks in the starting frames, these primitives are refined jointly with the static background during the propagation process. This streamlined strategy effectively captures complex shadow dynamics while maintaining the overall simplicity of the pipeline (\cf Sec.~\ref{sec:exp:ablation}).


%% file: sec/4_experiments.tex
\begin{figure*}[t]
    \centering
    \includegraphics[width=0.99\linewidth]{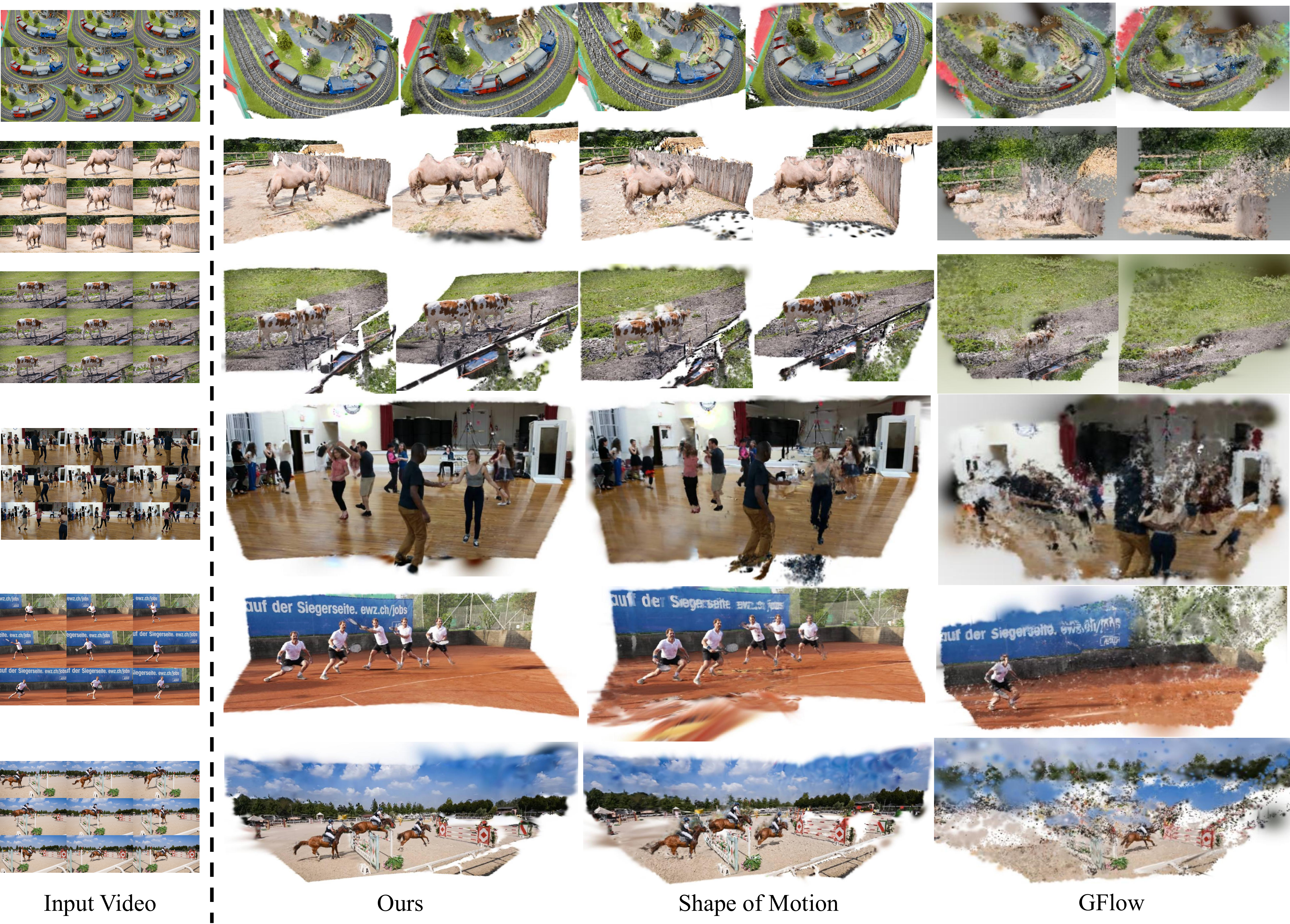}
    \caption{Comparison of dynamic scene reconstruction results on challenging real-world videos from DAVIS dataset. We compare MotionScale with Shape of Motion~\cite{wang2025shape} and GFlow~\cite{wang2025gflow} on several dynamic scenes containing complex object motions, occlusions, and large appearance variations. For the top rows, we show rendered results under two different viewpoints for each compared method.}
    \label{fig:davis}
\end{figure*}

\section{Experiments}
\label{sec:exp}

\subsection{Experimental Setup}
\label{sec:exp:setup}
\smallskip
\noindent\textbf{Implementation details.}
We implement MotionScale in PyTorch and conduct all experiments on a single NVIDIA RTX 4090 GPU. By default, for in-the-wild sequences, we utilize $\pi^3$~\cite{wang2025pi3} for monocular depth and camera poses, and SAM2~\cite{ravi2024sam} for foreground masks. 2D point tracks are generated by sampling a dense grid within these masks and tracking them via CoTracker3~\cite{karaev2025cotracker3}. 
To ensure fair comparison on established benchmarks (e.g., DyCheck), we strictly follow their provided evaluation protocols and utilize the standard input priors.
The scene is initialized with canonical 3D Gaussians following~\cite{kerbl20233d}. Optimization follows the progressive optimization strategy described in Sec.~\ref{sec:method}. Further details on hyperparameters and training schedules are provided in the supplementary material.

\smallskip
\noindent\textbf{Datasets.}
We evaluate our method on three diverse benchmarks: 
(1) \textit{DAVIS}~\cite{perazzi2016benchmark}, featuring single-view, casually recorded videos with complex non-rigid motions and natural camera movement; 
(2) \textit{DyCheck}~\cite{gao2022monocular}, containing 14 real-world dynamic scenes, including additional synchronized views from two static cameras and metric LiDAR depth. We utilize its sparse keypoint annotations (5-15 per sequence) to evaluate long-term 3D tracking; and 
(3) \textit{NVIDIA Dynamic Scenes}~\cite{yoon2020novel}, which provides calibrated multi-view sequences of diverse human and object activities. 
These datasets allow us to assess MotionScale's ability to recover consistent 4D representations from both casual monocular recordings and controlled multi-view rigs.

\smallskip
\noindent\textbf{Evaluation metrics.} 
For rendering and NVS, we report standard metrics: PSNR, SSIM, and LPIPS~\cite{zhang2018unreasonable}. 
For 3D tracking, we report the 3D End-Point-Error (EPE) and the percentage of points within distance thresholds $\delta_\text{3D}^{.05}$ and $\delta_\text{3D}^{.10}$ (5cm and 10cm). 
For 2D tracking, we follow standard protocols~\cite{doersch2022tap} and report Average Jaccard (AJ), average position accuracy ($\delta_\text{avg}$), and Occlusion Accuracy (OA).


\subsection{4D Scene Reconstruction}
\label{sec:exp:4d_recon}
We evaluate MotionScale on the monocular DAVIS benchmark~\cite{perazzi2016benchmark}, which presents significant challenges due to its single-view perspective and complex, non-rigid deformations.
We compare our approach against state-of-the-art dynamic reconstruction methods, including Shape-of-Motion~\cite{wang2025shape} and GFlow~\cite{wang2025gflow}.
As illustrated in the top portion of Fig.~\ref{fig:davis}, we render dynamic objects from two novel viewpoints across multiple timestamps to assess geometric and temporal consistency.
While GFlow frequently produces ``cloud-like'' artifacts and Shape-of-Motion exhibits visible geometric drift and temporal flickering, our cluster-based motion field effectively constrain Gaussian trajectories, preserving sharp surfaces and textural details.
Furthermore, in large-scale scenes characterized by extensive camera movement and rapid object dynamics (Fig.~\ref{fig:davis}, bottom), competing methods struggle with motion discontinuities during large displacements. Conversely, our progressive optimization strategy ensures a stable global structure and detailed local motion by grounding the motion field in the established background geometry. 
Overall, these results demonstrate that MotionScale achieves superior 4D fidelity, delivering clearer geometry and smoother motion across diverse real-world scenarios.

\begin{table}
    \centering
    \footnotesize
    \setlength{\tabcolsep}{1.5pt}
    \caption{Comparison of novel view synthesis results on DyCheck~\cite{gao2022monocular} and NVIDIA~\cite{yoon2020novel} datasets.}
    \begin{tabular}{l|ccc|ccc}
        \toprule
        \multirow{2}{*}{\textbf{Method}} & \multicolumn{3}{c|}{\textbf{DyCheck NVS}} & \multicolumn{3}{c}{\textbf{NVIDIA NVS}} \\
                                & PSNR$\uparrow$ & SSIM$\uparrow$ & LPIPS$\downarrow$ & PSNR$\uparrow$ & SSIM$\uparrow$ & LPIPS$\downarrow$ \\
        \midrule
        T-NeRF~\cite{gao2022monocular} & 15.60 & 0.55 & 0.55 & 20.76 & 0.59 & 0.17 \\
        HyperNeRF~\cite{park2021hypernerf} & 15.99 & 0.59 & 0.51 & 20.05 & 0.57 & 0.18 \\
        DynIBaR~\cite{li2023dynibar} & 13.41 & 0.48 & 0.55 & - & - & - \\
        Deformable-GS~\cite{yang2024deformable} & 11.92 & 0.49 & 0.66 & - & - & - \\
        4D Gaussians~\cite{wu20244d} & 13.11 & 0.39 & 0.73 & 17.69 & 0.48 & 0.38 \\
        Dynamic GM~\cite{stearns2024dynamic} & 15.79 & 0.59 & 0.44 & 22.36 & 0.66 & 0.15 \\
        4D-Fly~\cite{wu20254d} & 17.03 & 0.60 & 0.37 & 22.52 & 0.69 & 0.14 \\
        Shape of Motion~\cite{wang2025shape} & 16.72 & 0.63 & 0.45 & 23.37 & 0.75 & 0.10 \\
        \midrule
        MotionScale & \textbf{17.98} & \textbf{0.70} & 0.40 & \textbf{26.75} & \textbf{0.78} & \textbf{0.07} \\
        \bottomrule
    \end{tabular}
    \label{tab:nvs}
\end{table}

\subsection{Novel View Synthesis}
\label{sec:exp:novel}
We evaluate novel view synthesis (NVS) performance on the DyCheck~\cite{gao2022monocular} and NVIDIA Dynamic Scenes~\cite{yoon2020novel} benchmarks. As established in Sec.~\ref{sec:exp:setup}, we strictly follow the standard evaluation protocols and utilize provided input priors to ensure a direct comparison with existing baselines. Quantitative results are summarized in Tab.~\ref{tab:nvs}, where MotionScale consistently outperforms all prior methods on both datasets. On the DyCheck benchmark, our method's substantial gains in PSNR and lower LPIPS scores reflect superior photometric and perceptual fidelity. These improvements are particularly pronounced in dynamic regions involving large non-rigid motions, where competing approaches frequently produce blur or ghosting artifacts. Similarly, on NVIDIA Dynamic Scenes, MotionScale effectively preserves fine-grained motion details and maintains strict temporal coherence across synthesized views. These results validate that our proposed method successfully resolves complex temporal dynamics, delivering high-quality novel view synthesis in both casual monocular and calibrated multi-view settings.

\begin{table}[t]
    \centering
    \footnotesize
    \setlength{\tabcolsep}{2pt}
    \caption{Comparison of point-based tracking performance on the DyCheck~\cite{gao2022monocular} dataset.}
    \begin{tabular*}{\columnwidth}{@{\extracolsep{\fill}} l|ccc|ccc}
        \toprule
        \multirow{2}{*}{\textbf{Method}} &
        \multicolumn{3}{c|}{\textbf{3D Point Tracking}} &
        \multicolumn{3}{c}{\textbf{2D Point Tracking}} \\
        & EPE $\downarrow$ & $\delta_\text{3D}^{.05}\uparrow$ & $\delta_\text{3D}^{.10}\uparrow$
        & AJ $\uparrow$ & $<\delta_\text{avg}\uparrow$ & OA $\uparrow$ \\
        \midrule
        HyperNeRF~\cite{park2021hypernerf}                           & 0.182 & 28.4 & 45.8 & 10.1 & 19.3 & 52.0 \\
        DynIBaR~\cite{li2023dynibar}                                 & 0.252 & 11.4 & 24.6 & 5.4  & 8.7  & 37.7 \\
        Deformable-GS~\cite{yang2024deformable}                   & 0.151 & 33.4 & 55.3 & 14.0 & 20.9 & 63.9 \\
        DynMF~\cite{kratimenos2024dynmf}                             & 0.188 & 22.9 & 53.8 & 5.5  & 9.5  & 60.5 \\
        CoTracker~\cite{karaev2024cotracker}+DA~\cite{yang2024depth} & 0.202 & 34.3 & 57.9 & 24.1 & 33.9 & 73.0 \\
        TAPIR~\cite{doersch2023tapir}+DA~\cite{yang2024depth}        & 0.114 & 38.1 & 63.2 & 27.8 & 41.5 & 67.4 \\
        DELTA~\cite{ngo2024delta}                                    & 0.159 & 32.5 & 55.3 & 24.7 & 34.1 & 68.9 \\
        SpatialTracker~\cite{xiao2024spatialtracker}                 & 0.125 & 37.7 & 63.9 & 24.9 & 36.9 & 73.5 \\
        Shape of Motion~\cite{wang2025shape}                         & 0.082 & 43.0 & 73.3 & 34.4 & 47.0 & 86.6 \\
        \midrule
        MotionScale                                                  & \textbf{0.070} & \textbf{47.0} & \textbf{76.4}
                                                                     & \textbf{37.7} & \textbf{50.6} & \textbf{87.4} \\
        \bottomrule
    \end{tabular*}
    \label{tab:tracking}
\end{table}

\subsection{3D Point Tracking}
\label{sec:exp:tracking}
We evaluate the accuracy of our recovered motion field by performing point-based tracking on the DyCheck benchmark~\cite{gao2022monocular}. We evaluate 3D trajectories by lifting sparse 2D keypoints using LiDAR depth, alongside standard 2D tracking metrics. Quantitative results in Tab.~\ref{tab:tracking} show that MotionScale significantly outperforms existing dynamic reconstruction baselines across all metrics. For 3D tracking, our method achieves the lowest End-Point Error (EPE) and the highest accuracy under both $\delta_\text{3D}^{.05}$ and $\delta_\text{3D}^{.10}$ thresholds, demonstrating a robust ability to maintain 3D correspondences despite rapid non-rigid deformations. In the 2D evaluation, MotionScale attains notable gains in Average Jaccard (AJ) and Occlusion Accuracy (OA). These improvements suggest that our tracking loss and cluster-based motion propagation successfully resolve long-term dependencies and visibility changes, producing more stable and temporally coherent trajectories than methods relying on global or limited motion representations.

\begin{table}
    \centering
    \footnotesize
    \setlength{\tabcolsep}{2pt}
    \caption{Ablation studies on the DyCheck dataset.}
    \begin{tabular}{l|cccccc}
        \toprule
        \textbf{Method} & PSNR$\uparrow$ & SSIM$\uparrow$ & LPIPS$\downarrow$ & AJ$\uparrow$ & $<\delta_\text{avg}\uparrow$ & OA$\uparrow$ \\
        \midrule
        Ours (Full)             & \textbf{17.98} & \textbf{0.70} & \textbf{0.40} & \textbf{37.7} & \textbf{50.6} & \textbf{87.4} \\
        Global Bases            & 16.70 & 0.63 & 0.45 & 34.2 & 46.6 & 86.1 \\
        w/o Adaptive Control    & 17.21 & 0.67 & 0.42 & 34.9 & 47.0 & 86.6 \\
        w/o Pose Ref.           & 17.45 & 0.67 & 0.41 & - & - & - \\
        w/o Shadow              & 16.26 & 0.60 & 0.50 & - & - & - \\
        w/o FG Propagation      & 16.97 & 0.64 & 0.42 & 34.4 & 46.9 & 86.4 \\
        \bottomrule
    \end{tabular}
    \label{tab:ablation}
\end{table}

\subsection{Ablation Studies}
\label{sec:exp:ablation}
We conduct ablation experiments on the DyCheck benchmark~\cite{gao2022monocular} to evaluate the contribution of our key framework components. Quantitative results for NVS and 2D tracking are summarized in Tab.~\ref{tab:ablation}.

\smallskip
\noindent\textbf{Scalable motion field.}
To verify our motion representation, we compare against a baseline using a fixed number of global motion bases shared by all Gaussians (\textit{Global Bases}), similar to~\cite{wang2025shape}. Our cluster-based design significantly outperforms this baseline, demonstrating that localizing motion bases to specific Gaussian clusters provides the necessary degrees of freedom to capture fine-grained, non-rigid deformations that global bases tend to over-smooth.

\smallskip
\noindent\textbf{Adaptive Control.}
We ablate our dynamic cluster management by disabling the splitting and culling mechanisms (\textit{w/o Adaptive Control}). In this variant, cluster assignments remain fixed to the initialization. The performance drop indicates that the ability to topologically adapt the motion field and develop divergent dynamics is crucial for maintaining motion accuracy as the scene complexity evolves.

\begin{figure}[t]
    \centering
    \def\subsubfigwidth{0.3\linewidth}
    \begin{subfigure}{\subsubfigwidth}
        \centering
        \includegraphics[width=\linewidth]{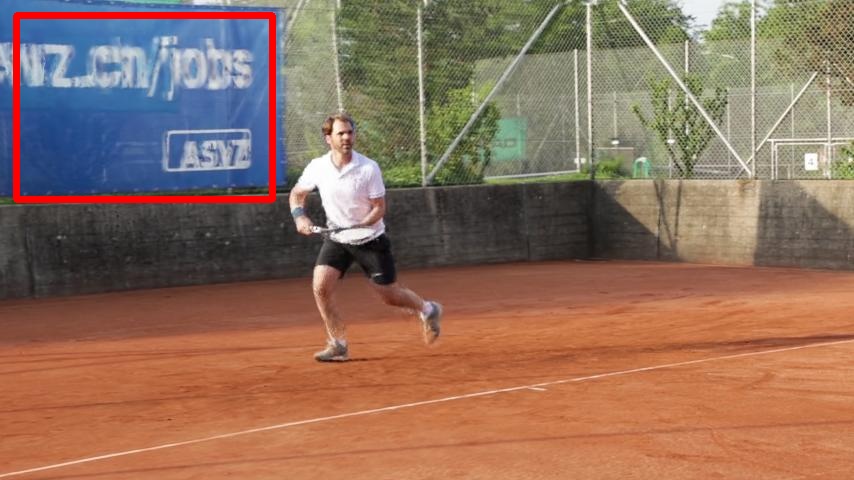}
        \caption{w/o Pose Ref.}
    \end{subfigure}%
    \begin{subfigure}{\subsubfigwidth}
        \centering
        \includegraphics[width=\linewidth]{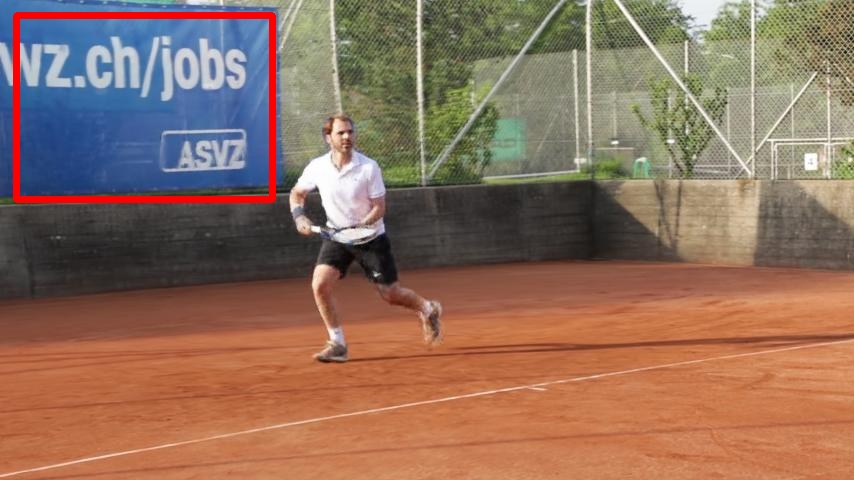}
        \caption{w/ Pose Ref.}
    \end{subfigure}%
    \begin{subfigure}{\subsubfigwidth}
        \centering
        \includegraphics[width=\linewidth]{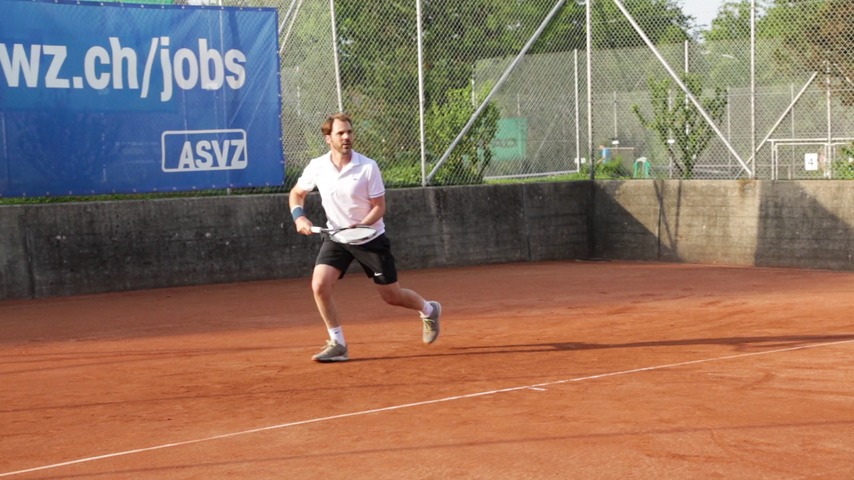}
        \caption{GT Image}
    \end{subfigure}

    \begin{subfigure}{\subsubfigwidth}
        \centering
        \includegraphics[width=\linewidth]{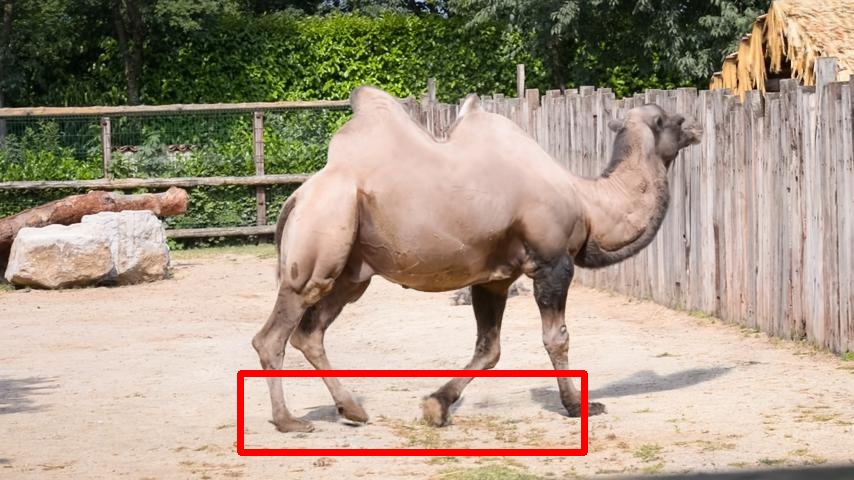}
        \caption{w/o Shadow}
    \end{subfigure}%
    \begin{subfigure}{\subsubfigwidth}
        \centering
        \includegraphics[width=\linewidth]{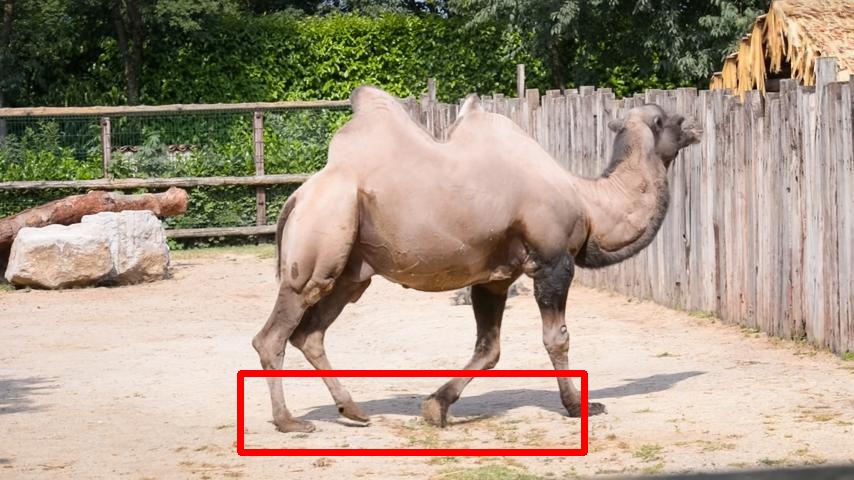}
        \caption{w/ Shadow}
    \end{subfigure}%
    \begin{subfigure}{\subsubfigwidth}
        \centering
        \includegraphics[width=\linewidth]{fig/shadow/00002.jpg}
        \caption{GT Image}
    \end{subfigure}
    
    \caption{Visual comparison of ablation results.}
    \label{fig:ablation}
\end{figure}

\smallskip
\noindent\textbf{Pose Refinement.} 
We evaluate the impact of joint pose refinement by fixing camera trajectories to initial $\pi^3$~\cite{wang2025pi3} estimates (\textit{w/o Pose Ref.}). As shown qualitatively in Fig.~\ref{fig:ablation} (top), this leads to degraded photometric alignment. The subtle geometric drift causes noticeable blurring in sharp textures (red box), confirming that high-quality monocular priors require spatial refinement during Gaussian optimization to ensure strict consistency.

\smallskip
\noindent\textbf{Shadow Gaussians.} 
We further ablate the role of our dedicated shadow primitives by disabling this subset in the background model (\textit{w/o Shadow}). Quantitatively, this leads to a significant decrease in PSNR for background rendering (Tab.~\ref{tab:ablation}). 
Visually, the absence of these primitives prevents the model from accurately reconstructing transient lighting on the ground (\cf Fig.~\ref{fig:ablation}, bottom).
More critically, without a dedicated representation, the optimization frequently forces foreground Gaussians to over-extend to capture shadow regions. 
This leads to unintended geometric dilation and ghosting artifacts on the dynamic objects themselves. Our design effectively decouples foreground geometry from transient shadows, ensuring both clean background textures and high-fidelity object reconstruction.

\smallskip
\noindent\textbf{Foreground Propagation.}
Finally, we assess our foreground propagation by replacing it with a global optimization approach initialized from full-frame tracks (\textit{w/o FG Propagation}). This variant fails to maintain temporal coherence over long sequences and is prone to optimization instability. This highlights that our three-stage refinement is essential for scaling 4D reconstruction to complex videos.

\section{Conclusion}
\label{sec:conclusion}
In this paper, we introduced \textbf{MotionScale}, a scalable framework designed to recover high-fidelity 4D representations from casual monocular video. While existing methods often achieve impressive view-dependent rendering, they frequently struggle to maintain geometric and temporal consistency over long sequences due to their reliance on short-horizon 2D priors and rigid motion representations. We addressed these challenges by proposing an adaptive, cluster-based motion field that dynamically adjusts its capacity through splitting and culling. Coupled with a progressive optimization strategy, MotionScale effectively bridges the gap between noisy 2D tracks and coherent 3D geometry by incrementally incorporating temporal context. Our comprehensive evaluation across the DAVIS, DyCheck, and NVIDIA benchmarks demonstrates that MotionScale significantly outperforms state-of-the-art approaches in rendering quality, motion accuracy, and geometric stability. 
